\documentclass{amsart}

\usepackage{amsmath}
\usepackage{amsthm}
\usepackage{amssymb}
\usepackage[dvips]{graphicx}
\usepackage{algorithm}
\usepackage{algorithmic}

\newtheorem{theorem}{Theorem}
\newtheorem{lemma}[theorem]{Lemma}
\newtheorem{definition}[theorem]{Definition}
\newtheorem{corollary}[theorem]{Corollary}
\newtheorem{proposition}[theorem]{Proposition} 
\newcommand{\BlackBox}{\rule{1.5ex}{1.5ex}}  
\renewenvironment{proof}{\par\noindent{\bf Proof\ }}{\hfill\BlackBox\\[2mm]}

\begin{document} 
 
\begin{center}
{\Large \bf GroupLiNGAM: Linear non-Gaussian acyclic models\\for sets of variables}\\
\vspace{5mm}
Yoshinobu Kawahara,\footnote{The Institute of Scientific and Industrial Research (ISIR), Osaka University.\\
\hspace{6mm}E-mail: {\texttt\{kawahara,sshimizu,washio\}@ar.sanken.osaka-u.ac.jp}}
Kenneth Bollen,\footnote{Department of Sociology, The University of North Carolina.\\
\hspace{6mm}E-mail: {\texttt bollen@unc.edu}}
Shohei Shimizu$^\text{1}$ and
Takashi Washio$^\text{1}$
\vspace{8mm}
\begin{quote}
{\small {\bf Abstract:} Finding the structure of a graphical model has been received much attention in many fields.~Recently, it is reported that the non-Gaussianity of data enables us to identify the structure of a directed acyclic graph without any prior knowledge on the structure.~In this paper, we propose a novel non-Gaussianity based algorithm for more general type of models; chain graphs.~The algorithm finds an ordering of the disjoint subsets of variables by iteratively evaluating the independence between the variable subset and the residuals when the remaining variables are regressed on those.~However, its computational cost grows exponentially according to the number of variables.~Therefore, we further discuss an efficient approximate approach for applying the algorithm to large sized graphs.~We illustrate the algorithm with artificial and real-world datasets.}
\end{quote}
\vspace{5mm}
\end{center}

\section{Introduction}
\label{se:intro}
 
In order to discover or understand data generating mechanisms, a graphical model has been used as a fundamental tool, and the problem of finding its structure from data has been received much attention in many fields including social sciences \cite{Bol89}, bioinformatics \cite{RS07} and neuroinformatics \cite{LDBB06}.

Among variety of models, structural equation models (SEMs) and Bayesian networks (BNs) have been widely used to analyze causal relationships in empirical studies \cite{Bol89,Pea00,SGS01}. However, the full structure, {\em i.e.}, a causal ordering and connection strengths, of the model cannot be identified in most cases without prior knowledge on the structure when only covariance structure of data is used for model estimation as is the case in almost conventional methods. Recently, it is reported that non-Gaussian structure of data overcomes this identifiability problem in the case of linear directed acyclic graphs (DAGs) \cite{SHHK06,SHKW09}. By their algorithms (the LiNGAM algorithms), if the external influences are non-Gaussian, the structure can be uniquely estimated by using observed data only without any prior knowledge (under an assumption of acyclicity).

However, the applicability of the LiNGAM algorithms might be restricted in some real-world applications because of its relatively strong assumptions of linear acyclicity in each variable. For example, in the cases where an unobserved confounder exists between exogenous variables or sink variables, a DAG structure is no longer appropriate to apply. Thus, it would be useful to develop a non-Gaussianity based framework to estimate the structure of more general class of models, such as chain graphs \cite{Lau96}, so as to deal with the situations under which the assumptions on DAGs are not satisfied. Note that there is a method based on non-Gaussianity that takes unobserved confounders into account \cite{HSKP08}. However, since this method needs to model unobserved variables explicitly, the computational cost is crucially high (In fact, only two or three variables were empirically treated in the paper).

In this paper, we propose a non-Gaussian variant of chain graphs, which includes the one of linear acyclic graphs as a special case, and present an algorithm for the estimation of this model. The algorithm finds an ordering of the subsets of variables by iteratively evaluating the independence between the variable subset and the residuals when the remaining variables are regressed on those. In addition to the applicability to chain
graphs, it is empirically verified that the estimation by the proposed algorithm works reasonably well compared with the existing algorithms when applied to DAGs. However, this procedure needs to compute the independence exponentially many times corresponding to the number of variables. Therefore, we propose an approximate approach that can be performed without depending on the number of variables (although the accuracy may depend on) and can be applied to large scale graphs. The performance will be illustrated using artificial and real-world datasets.

The reminder of this paper is organized as follows.
In Sect.~\ref{se:glingam}, we first introduce a linear non-Gaussian acyclic model for sets of variables (GroupLiNGAM model). 
Then in Sect.~\ref{se:est_glingam}, we present an algorithm for (directly) estimating the GroupLiNGAM model. However, this approach would be inefficient for large sized graphs. Therefore, in Sect.~\ref{se:approx}, we give an approximate approach that can be applied to large sized graphs based on the algorithm described in Sect.~\ref{se:est_glingam}. The algorithms are illustrated and examined in its performance using artificial data in Sect.~\ref{se:sim} and real-world data in Sect.~\ref{se:real}. Finally, we give conclusions in Sect.~\ref{se:conclusion}.

\section{GroupLiNGAM model}
\label{se:glingam}
In this paper, we consider a non-Gaussian variant of chain graphs, which we call the {\em GroupLiNGAM model}.

Assume that observed data are generated from a process represented graphically by a chain graph on random variables $\boldsymbol{x}$ of dimension $p$. Let us express this chain graph by a $p\times p$ adjacency matrix $B=\{b_{ij}\}$, where every $b_{ij}$ represents the connection strength from a variable $x_j$ to another $x_i$ in the chain graph. Also, let $K(l)$ ($l=1,\ldots,m$, $m\leq p$) be ordered blocks, {\em i.e.}, disjoint subsets of variables, so that no variables in later subsets influence any variable in earlier subsets and $K(1)\cup\cdots\cup K(m)=V$, where $V:=\{1,\ldots,p\}$ is the indices set of the variables.\footnote{This definition is a generalization of the one of a DAG. That is, this is actually the definition of a DAG if all the subsets consist of one element, {\em i.e.},  $m=p$.} The index of the subset, {\em i.e.}, $l$, that $x_i$ belongs to will be referred as $l(i)$. Moreover, assume that the relations between variables in different subsets are linear. Without loss of generality, each observed variable $x_i$ is assumed to have zero mean.

Then, the GroupLiNGAM model is represented as
\begin{equation}
\label{eq:glingam1}
x_i = \sum_{l(j)\leq l(i),i\neq j} b_{ij} x_j + e_i,
\end{equation}
where $e_i$ is an external influence. All external influences $e_i$'s are non-Gaussian random variables with zero means and non-zeros variances, and independent of each other in different blocks. Alternatively, we write the model \eqref{eq:glingam1} in a matrix form:
\begin{equation}
\label{eq:glingam2}
\boldsymbol{x} = B\boldsymbol{x}+\boldsymbol{e},
\end{equation}
where $B$ can be permuted by simultaneous equal row and column permutations to be lower block-triangular due to the acyclicity of disjoint subsets in chain graphs \cite{WL90,CW96}. Moreover, if we represent the model \eqref{eq:glingam2} as
\begin{equation}
\label{eq:glingam3}
\boldsymbol{x} = A\boldsymbol{e},
\end{equation}
the matrix $A~(:=(I-B)^{-1})$ (called a mixing matrix) also becomes lower block-triangular (and with all unities in the diagonal). Note that, in the case of $m=p$, {\em i.e.}, the DAG case, the model \eqref{eq:glingam3} defines the independent component analysis (ICA) model \cite{HKO01} since the components of $\boldsymbol{e}$ are independent and non-Gaussian. Since the ICA model is identifiable, the model \eqref{eq:glingam3} in this case ($m=p$) is also identifiable, which is the key idea of the original LiNGAM algorithm \cite{SHHK06} (we call it ICA-LiNGAM in the later part of this paper).

\begin{figure}[t]
\centering
\includegraphics[keepaspectratio=true,width=.75\linewidth]{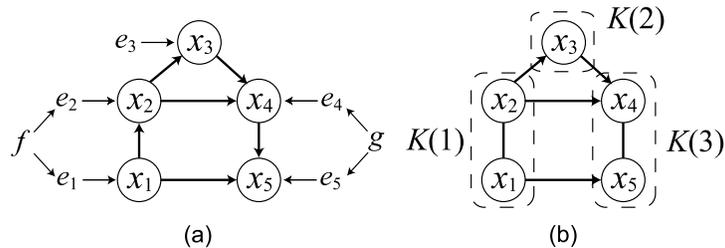}
\caption{Illustrative example of a chain graph.}
\label{fig:example}
\end{figure}

Now, let us consider an illustrative example in which the model is represented by (cf.~Figure~\ref{fig:example}~(a))
\begin{equation}
\label{eq:example}
\begin{split}
x_1 &= e_1,\\
x_2 &= b_{21}x_1 + e_2,\\
x_3 &= b_{32}x_2 + e_3,\\
x_4 &= b_{42}x_2 + b_{43}x_3 + e_4,\\
x_5 &= b_{51}x_1 + b_{54}x_4 + e_5,
\end{split}
\end{equation}
where unobserved confounders $f$ and $g$ exist between $e_1$ and $e_2$ and between $e_4$ and $e_5$, respectively, as
\begin{equation*}
e_1 = c_1 f + d_1
\hspace{2mm}\text{and}\hspace{2mm}
e_2 = c_2 f + d_2,
\end{equation*}
and
\begin{equation*}
e_4 = c_4 g + d_4
\hspace{2mm}\text{and}\hspace{2mm}
e_5 = c_5 g + d_5.
\end{equation*}
$d_1$, $d_2$, $d_4$ and $d_5$ are independent of each other. Note that, in this case, the assumption for the LiNGAM algorithms, {\em i.e.}, exogenous influences are independent of each other, is not satisfied. In fact, $x_1$ and $x_4$ depend on $x_2$ and $x_5$, respectively, because $f$ and $g$ are not observed, and a DAG representation is no longer appropriate to apply. The ordered blocks for the example \eqref{eq:example} are $K(1)=\{1,2\}$, $K(2)=\{3\}$ and $K(3)=\{4,5\}$ (cf.~Figure~\ref{fig:example}~(b)).

\section{Model estimation}
\label{se:est_glingam}

In this section, we address the estimation of the GroupLiNGAM model from data. In the following parts, we will refer the subset of variables corresponding to $S\subseteq V$ as $\boldsymbol{x}_S$. Also, we denote by $V\setminus S$ the complementary set of $V$ with respect to $S$, and by $\boldsymbol{x}_{\bar{S}}$ the subset of variables corresponding to $V\setminus S$.

\subsection{Identifying exogenous variables using non-Gaussianity}
\label{ss:exogenous}

Recently, it has been reported that non-Gaussianity of external influences serves for directly estimating the ordering of variables from data \cite{SHKW09} (DirectLiNGAM). The key insight herein is that, once an exogenous variable is identified, we can remove the component of the exogenous variable from the other variables without violating the original ordering for the residuals when the exogenous variable is regressed on the remaining variables. Here, we describe the analogous insight still holds for sets of variables. To this end, we first need the following assumption:
\begin{definition}[correlation-faithfulness]
The distribution of $\boldsymbol{x}$ is said to be correlation-faithful to the generating graph if correlation and conditional correlation of $x_i$ are entailed by the graph structure, i.e., the zeros/non-zeros status of $b_{ij}$, but not by specific parameter values of $b_{ij}$.
\end{definition}
This concept is motivated by the faithfulness \cite{SGS01}. Also, we give the definition of the exogenous set of variables as follows.
\begin{definition}[exogenous set]
\label{def:exogenous}
Let the partition of the variables $\boldsymbol{x}$ be $\boldsymbol{x}=(\boldsymbol{x}_S,\boldsymbol{x}_{\bar{S}})$ such that $\boldsymbol{x}_S$ and $\boldsymbol{x}_{\bar{S}}$ are not empty. Then, the subset of variables $\boldsymbol{x}_S$ is said to be exogenous against $\boldsymbol{x}_{\bar{S}}$, if the corresponding partition of the the matrix $B$ has the following form:
\begin{equation*}
B = \left[\begin{array}{cc}B_{S}&0\\B_{\bar{S},S}&B_{\bar{S}}\end{array}\right].
\end{equation*}
\end{definition}
Note that each variable in the exogenous set is not necessarily an exogenous variable. That is, the variables in an exogenous set may be influenced by each other inside of the set. Also note that the submatrix of the mixing matrix $A$ corresponding to $B_S$ is full-rank because the covariance matrix $\Sigma_S$ of $\boldsymbol{x}_S$ is also full-rank from the correlation-faithfulness assumption.

Now, we give two lemmas and one corollary that is the basis of the algorithm proposed in this paper.
\begin{lemma}
\label{le:exogenous}
Assume that the input data $\mathbf{x}$ follows the GroupLiNGAM model \eqref{eq:glingam2}, and that the distribution of $\boldsymbol{x}$ is correlation-faithful to the generating graph. Let $r^{(S)}$ be the residual when $x_{\bar{S}}$ is regressed on $\boldsymbol{x}_S$ for $S\subset V$ {\em :} $r^{(S)}=x_{\bar{S}}-\Sigma_{S,\bar{S}}^T\Sigma_S^{-1}\boldsymbol{x}_S$, where
\begin{equation*}
\Sigma = \left[\begin{array}{cc}
           \Sigma_S      & \Sigma_{S,\bar{S}} \\
           \Sigma_{S,\bar{S}}^T & \Sigma_{\bar{S}}
         \end{array}\right]
\end{equation*}
is the covariance matrix of $(\boldsymbol{x}_S,\boldsymbol{x}_{\bar{S}})$. Then, a set of variables $\boldsymbol{x}_S$ is exogenous if and only if $\boldsymbol{x}_S$ is independent of its residual $\boldsymbol{r}^{(S)}$.
\end{lemma}
\begin{proof}
First, assume that $\boldsymbol{x}_S$ is exogenous. Then, one can write $\boldsymbol{x}_{\bar{S}}=A_{\bar{S},S}A_S^{-1}\boldsymbol{x}_S + \bar{\boldsymbol{e}}_{\bar{S}}^{(S)}$, where
\begin{equation*}
A = \left[\begin{array}{cc}A_S&0\\A_{\bar{S},S}&A_{\bar{S}}\end{array}\right]
\end{equation*}
is the coefficient matrix in Eq.~\eqref{eq:glingam3}. From the definition of model~\eqref{eq:glingam3}, $\bar{\boldsymbol{e}}_{\bar{S}}^{(S)}=A_{\bar{S}}\boldsymbol{e}_{\bar{S}}$ and $\boldsymbol{x}_S$ are mutually independent. Also, since $\Sigma_{S,\bar{S}}^T=A_{\bar{S},S}A_S^{-1}\Sigma_S$, $A_{\bar{S},S}A_S^{-1}$ is equivalent to the regression coefficients when $\boldsymbol{x}_{\bar{S}}$ is regressed on $\boldsymbol{x}_S$. Therefore, $\boldsymbol{r}^{(S)}$ is equivalent to $\bar{\boldsymbol{e}}_{\bar{S}}^{(S)}$. As a result, $\boldsymbol{x}_S$ and $\boldsymbol{r}^{(S)}$ are mutually independent.
Next, assume that $\boldsymbol{x}_S$ is independent of $\boldsymbol{r}^{(S)}$. Then, since $\boldsymbol{x}_S$ is independent of $\boldsymbol{e}_{\bar{S}}$, all elements of the regression coefficient matrix when $\boldsymbol{x}_S$ is regressed on $\boldsymbol{e}_{\bar{S}}$, {\em i.e.}, $A_{S,\bar{S}}$, become zeros, which means all elements of the upper right part of $B$, {\em i.e.}, $B_{S,\bar{S}}$, are also zeros. From the correlation-faithfulness assumption and  the definition of exogeneous sets, $\boldsymbol{x}_S$ is exogenous.
\end{proof}
\begin{lemma}
\label{le:residual}
Assume the assumptions of Lemma~\ref{le:exogenous} and that a set of variables $\boldsymbol{x}_S$ is exogenous. Let $\boldsymbol{r}^{(S)}$ be the residual vector when $\boldsymbol{x}_{\bar{S}}$ is regressed on $\boldsymbol{x}_S$ for $S\subset V$. Then, GroupLiNGAM models hold both for $\boldsymbol{x}_S$ and $\boldsymbol{r}^{(S)}$, respectively {\em :} $\boldsymbol{x}_S=B_S\boldsymbol{x}_S+\boldsymbol{e}_S$ and $\boldsymbol{r}^{(S)}=B^{(S)}\boldsymbol{r}^{(S)}+\boldsymbol{e}^{(S)}$, where $B_S$ and $B^{(S)}$ are matrices that can be permuted to be block lower-triangular by simultaneous row and column permutations, and elements of $\boldsymbol{e}_S$ and  $\boldsymbol{e}^{(S)}$ are non-Gaussian and mutually independent in different blocks, respectively.
\end{lemma}
\begin{proof}
Without loss of generality, assume that $B$ in the GroupLiNGAM model \eqref{eq:glingam2} is already permuted to be lower block-triangular (which means $A$ is also to be lower block-triangular with all unities in the diagonal). First, it is straightforward from Def.~\ref{def:exogenous} that $B_S$ is lower block-triangular. Next, since $\boldsymbol{x}_S$ is exogenous, the regression coefficients when $\boldsymbol{x}_{\bar{S}}$ is regressed on $\boldsymbol{x}_S$ becomes $A_{\bar{S},S}A_S^{-1}$. Therefore, removing the effects of $\boldsymbol{x}_S$ from $\boldsymbol{x}_{\bar{S}}$ by least-squares estimation is equivalent to setting all elements of the first $|S|$ columns of $A$ to be zeros. This means that the residuals $\boldsymbol{r}^{(S)}$ are not influenced by $\boldsymbol{x}_S$ because of the correlation-faithfulness assumption. As a result, we again obtain a lower block-triangular mixing matrix with all unities in the diagonal $A^{(S)}(=A_{\bar{S}})$ for $\boldsymbol{r}^{(S)}$.
\end{proof}
\begin{corollary}
\label{co:order}
Assume the assumptions in Lemma~\ref{le:residual}. Denote by $l_S(i)$ and $l_{\boldsymbol{r}^{(S)}}(i)$ the indices of the ordered subsets encoded by the chain graphs on $\boldsymbol{x}_S$ and $\boldsymbol{r}^{(S)}$, respectively. Recall that $l(i)$ denotes the index of the ordered subsets encoded by the chain graph on $\boldsymbol{x}$. Then, the ordering of the subsets of $\boldsymbol{x}_S$ and $\boldsymbol{r}^{(S)}$ are respectively equivalent to that of corresponding original subsets of variables, i.e., $l_S(i_1)<l_S(i_2)\Leftrightarrow l(i_1)<l(i_2)$ and $l_{\boldsymbol{r}^{(S)}}(i_1)<l_{\boldsymbol{r}^{(S)}}(i_2)\Leftrightarrow l(i_1)<l(i_2)$.
\end{corollary}
\begin{proof}
As described in the proof of Lemma~\ref{le:residual}, the adjacency matrices (and the mixing matrices) for the GroupLiNGAM models on $\boldsymbol{x}_S$ and $\boldsymbol{r}^{(S)}$ are equivalent to the corresponding parts of the one for the GroupLiNGAM model on $\boldsymbol{x}$. This shows the orderings of $\boldsymbol{x}_S$ and $\boldsymbol{r}^{(S)}$ are not changed.
\end{proof}

\begin{algorithm}[t]
\caption{GroupLiNGAM}
\label{alg:glingam}
\begin{algorithmic}[1]
\STATE Given a $p$-dimensional variables $\boldsymbol{x}$, a set of its subscripts $V$, a $p\times n$ data matrix of the variables $\mathbf{X}$, initialize an ordered subset of variables as $K\leftarrow \emptyset$.
\STATE Call $K\leftarrow$ GroupSearch~($V$, $K$, $\mathbf{X}$).
\STATE Construct a lower block-triangular matrix $B$ by following the order in $K$, and estimate the connection strengths $b_{ij}$ (using some conventional covariance-based regression, such as least-squares and maximum-likelihood approaches) on the original variables $\boldsymbol{x}$ and data matrix $\boldsymbol{X}$.
\end{algorithmic}
\vspace{1mm}
\begin{flushleft}
function~$K \leftarrow$ GroupSearch~($U$, $K$, $\mathbf{X}_U$)
\end{flushleft}
\begin{algorithmic}[1]
\FOR{$S\subset U$}
  \STATE Perform least-squares regression of $\boldsymbol{x}_S$ on $\boldsymbol{x}_{U\setminus S}$ (denote the residual vector by $\boldsymbol{r}^{(S)}$ and its residual data matrix by $\mathbf{R}^{(S)}$) and then compute some independence measure $I(S)$ between $\boldsymbol{x}_S$ and $\boldsymbol{r}^{(S)}$, {\em e.g.}, $MI(\boldsymbol{x}_S,\boldsymbol{r}^{(S)})$.
\ENDFOR
\STATE $S_*:=\arg\min I(S)$.
\IF{$I(S_*)\leq \delta$ and $|U|\neq1$}
  \STATE Set $\mathbf{X}_{U\setminus S_*}\leftarrow \mathbf{R}^{(S_*)}$.
  \STATE Call $K\leftarrow$ GroupSearch~$(S_*,K,\mathbf{X}_{S_*})$
  \STATE Call $K\leftarrow$ GroupSearch~$(U\setminus S_*,K,\mathbf{X}_{U\setminus S_*})$
\ELSE
  \STATE Append $S_*$ to the end of $K$.
\ENDIF
\end{algorithmic}
\end{algorithm}

Lemma~\ref{le:exogenous} indicates that an exogenous set is identified by evaluating the independence between a set of variables $\boldsymbol{x}_S$ and its residuals $\boldsymbol{r}^{(S)}$. Lemma~\ref{le:residual} implies that the GroupLiNGAM models for the $p$-dimensional vector $\boldsymbol{x}_S$ and the $(p-|S|)$-dimensional residual vector $\boldsymbol{r}^{(S)}$ can be handled as new input models, and Lemma~\ref{le:exogenous} can be further applied to the each model to derive the next set of exogenous variables. This process can be repeated until all subsets of variables are not able to be devided, and the resulting order of the sets of variable subscripts shows the causal order of the original observed variables according to Corollary~\ref{co:order}.

As the independence measure used in Lemma~\ref{le:exogenous}, the mutual information between the subset of variable and the residuals, {\em i.e.}, $MI(\boldsymbol{x}_S,\boldsymbol{r}^{(S)})$, would be available. There are many options for its estimation from data. In the later experiments, we used an algorithm based on the $k$-nearest neighbors method \cite{KSG04}.\footnote{We used the MATLAB code available from \texttt{http://www.klab.caltech.edu/$\sim$kraskov/MILCA/} in the experiments.} This method has one tuning parameters, {\em i.e.}, the number of neighbors $kneig$. Although the setup of this parameter is not trivial, the algorithm is known to work well empirically when $kneig$ is set as 3--5 \% of the dimensional $p$ \cite{KSG04}.

\subsection{GroupLiNGAM algorithm}
\label{ss:glingam}
Based on the above result, we now present an algorithm to estimate a block causal ordering and the connection strengths in the GroupLiNGAM model under the correlation-faithfulness assumption. The pseudo-code of the algorithm is shown in Alg.~\ref{alg:glingam}.

\begin{figure}[t]
\centering
\includegraphics[keepaspectratio=true,width=.75\linewidth]{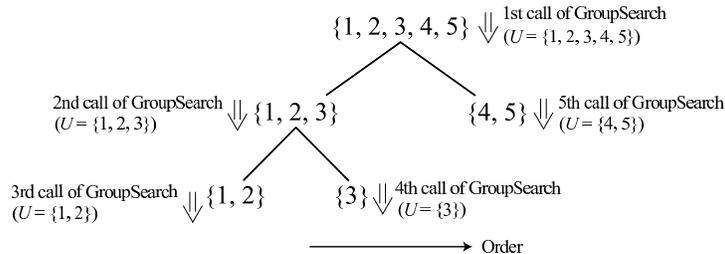}
\caption{Illustration of Alg.~\ref{alg:glingam} for the example \eqref{eq:example}.}
\label{fig:alg1}
\end{figure}

The algorithm is performed by the recursive calls of GroupSearch function, which devides a given subset $U$ into ordered two groups. Since an exogenous set is identified by evaluating the independence between a subset of $U$ and its residuals from Lemma~\ref{le:exogenous}, we find such subset $S_*(\subset U)$ as the one that minimizes some independence measure $I(S)$ (Lines 1--3 in GroupSearch in Alg.~\ref{alg:glingam}). Thus, $U$ is divided into ordered two groups $S_*$ and $U\setminus S_*$. From Lemma~\ref{le:residual}, for each of $S_*$ and $U\setminus S_*$, the GroupLiNGAM models hold. Therefore, this procedure is iterated until further partition cannot be found, which is judged with a threshold $\delta$ (Line~8--9 in GroupSearch in Alg.~\ref{alg:glingam}). Finally-obtained order of variable subsets is consistent globally, which is guaranteed by Corollary~\ref{co:order}. The illustration of this procedure for the example Eq.~\eqref{eq:example} is shown in Fig.~\ref{fig:alg1}.

Note that Alg.~\ref{alg:glingam} is specialized to the DAG case if we set $\delta=+\infty$. However, the outputs by Alg.~\ref{alg:glingam} and the DirectLiNGAM algorithm are not always same because Alg.~\ref{alg:glingam} finds the subset of variables that is exogenous against the remaining variables while the DirectLiNGAM algorithm identifies an exogenous variable iteratively (that is, we can say that the former uses global information  of independence between variables while the latter local one). Thus, the accumulation of errors of regression in Alg.~\ref{alg:glingam} is expected to be no more than the one in the DirectLiNGAM algorithm, which will be illustrated empirically in Sect.~\ref{se:sim}.

\section{Approximate approach for large graphs}
\label{se:approx}

Since Alg.~\ref{alg:glingam} needs to compute independence between $\boldsymbol{x}_S$ and $\boldsymbol{r}^{(S)}$ exponentially many times ($2^{|U|-1}$, once for every $S\subset U$)\footnote{$U=V$ ($|V|=p$) at the first iteration.} at each iteration (Lines~1--3 in GroupSearch), it can be applied only to medium sized graphs (consisting of up to around 15 nodes). Here, we propose an approximate approach based~on Alg.~\ref{alg:glingam} applicable to larger sized graphs (L-GroupLiNGAM).

The basic idea of the proposed algorithm is as follows. If we observe only a subset of variables, then some of the unobserved variables may act as confounders against some of the observed variables and, as a result, causal directions between such observed variables have become not identifiable. However, since the definition of our model permits confounders, {\em i.e.}, makes edges undirected if there exist confounders between the observed variables, we can find the order of blocks which is identifiable from the currently observed variables using Alg.~\ref{alg:glingam}. Therefore, by randomly picking up the subset of variables such that the set of the subsets covers all variables and applying Alg.~\ref{alg:glingam} to each subset, we finally obtain a block ordering of all variables in a large graph. The validness of this procedure can be guaranteed by the following proposition:
\begin{proposition}
Assume the assumptions in Lemma~\ref{le:residual}. Let denote by $\tilde{l}_T(i)$ ($T\subset V$) the ordering of the subsets of variables when only $\boldsymbol{x}_T$ is observed (and other variables $\boldsymbol{x}_{\bar{T}}$ are not observed). Then, the order $\tilde{l}_T(i)$ is consistent with the one when all variables are observed, i.e., $\tilde{l}_T(i_1)<\tilde{l}_T(i_2)\Rightarrow l(i_1)<l(i_2)$.
\end{proposition}
\begin{proof}
Assume that only $\boldsymbol{x}_T$ is observed for $T\subset V$. If $\tilde{l}_T(i_1)<\tilde{l}_T(i_2)$, then there exists a subset $S\subset T$ exogenous against $T\setminus S$ such that $i_1\in S$ and $i_2\in T\setminus S$. Therefore, one can write  $\boldsymbol{x}_S = \tilde{A}_S\tilde{\boldsymbol{e}}_S$ and $\boldsymbol{x}_{T\setminus S}=\tilde{A}_{T\setminus S,S}\tilde{A}_S^{-1}\boldsymbol{x}_S + \tilde{A}_{T\setminus S}\tilde{\boldsymbol{e}}_{T\setminus S}$, where $\boldsymbol{x}_S$ and $\tilde{A}_{T\setminus S}\tilde{\boldsymbol{e}}_{T\setminus S}$ are mutually independent. This means, if we denote as $\tilde{\boldsymbol{e}}_S=\sum_{i\in S_1} \boldsymbol{a}_{S,i}e_i$ and $\tilde{\boldsymbol{e}}_{T\setminus S}=\sum_{i\in S_2}\boldsymbol{a}_{T\setminus S,i}e_i$, where $S_1\subseteq S\cup (V\setminus T)$ and $S_2\subseteq (T\setminus S)\cup (V\setminus T)$, then the union of $S_1$ and $S_2$ is empty, {\em i.e.}, $S_1\cap S_2 = \emptyset$. This means that all elements of the submatrix $\{a_{ij}\}~(i\in S_1\cup S, j\in V\setminus (S_1\cup S))$ of the mixing matrix $A$ are zeros and, as a result, $l(i_1)<l(i_2)$.
\end{proof}

\begin{algorithm}[t]
\caption{L-GroupLiNGAM}
\label{alg:glingam2}
\begin{algorithmic}[1]
\STATE Given a $p$-dimensional variables $\boldsymbol{x}$, a set of its subscripts $V$, a $p\times n$ data matrix of the variables $\mathbf{X}$ and a cardinality $h$, initialize the list of orders between combinations of variables $\tilde{k}=\emptyset$.
\STATE Compute a random covering $T(i)~(i=1,\ldots,N)$ of variables with cardinarity $h$.
\FOR{$i=1,\ldots,N$}
\STATE Apply Alg.~\ref{alg:glingam} modefied by replacing Line 1 in GroupSearch func.\@ to \eqref{eq:line3} with $V\leftarrow T(i)$, and add new orders from its output $K$ to $\tilde{k}$.
\ENDFOR
\STATE Construct a block order $\tilde{K}$ for all variables from $\tilde{k}$.
\STATE Construct a strictly lower block-triangular matrix $B$ by following the order in $\tilde{K}$, and estimate the connection strengths $b_{ij}$ (using some conventional covariance-based regression, such as least-squares and maximum-likelihood approaches) on the original variables $\boldsymbol{x}$ and data matrix $\boldsymbol{X}$.
\end{algorithmic}
\end{algorithm}
Based on the above result, we now present an algorithm for estimating the GroupLiNGAM model with large number of variables. The pseudo-code of the algorithm is shown in Alg.~\ref{alg:glingam2}, where $\tilde{k}$ is the list of combinations of variables $(j_1,j_2)$ with orders $l(j_1)<l(j_2)$.

In the algorithm, we first generate a random covering of all variables $T(i)~(i=1,\ldots,N)$ (Line~2 in Alg.~\ref{alg:glingam2}), {\em i.e.}, subsets $T(i)\subset V$ such that $\cup_{i=1,\ldots,N} T(i)=V$. And, we apply Alg.~\ref{alg:glingam} to each $T(i)$ (Lines~3--5 in Alg.~\ref{alg:glingam2}). Then, in order to reflect already-known orders $(j_1,j_2)$ $(j_1,j_2\in T(i))$ when choosing $S\subset U$ in Lines~1--3 in GroupSearch function in Alg.~\ref{alg:glingam}, we replace Lines 1 in GroupSearch to the following:
\begin{equation}
\label{eq:line3}
\text{{\bf for} $S\subset U$ s.t.\@ $j_2\in S\rightarrow j_1\notin U\setminus S$ for $(j_1,j_2)\in \tilde{k}$ {\bf do}}
\end{equation}
Also, the application of Alg.~\ref{alg:glingam} may generate an output making a cycle as a whole when combined with previously-obtained orders due to statistical uncertainty of samples as the iterations of Lines 3--5 in Alg.~\ref{alg:glingam2} continue. In such a case, the inconsistent (old and new) orders need to be removed, {\em i.e.}, we merge the ordered variables by these orders into a group. Finally, the ordering of variable subsets is constructed from the list of obtained orders. Although this procedure may not be able to find some of block orders, more and more ones are expected to be found depending on subsets $T(i)$ as the iteration continue.

\section{Simulations}
\label{se:sim}

In this section, we evaluate the proposed algorithms empirically using artificial datasets. Especially, we focus on (i) the evaluation of the validity of the proposed algorithms for estimating the GroupLiNGAM model (Alg.~\ref{alg:glingam} and Alg.~\ref{alg:glingam2}) and (ii) the comparison of estimation accuracy of the proposed and existing algorithms (ICA-LiNGAM \cite{SHHK06} and DirectLiNGAM \cite{SHKW09}) in DAG cases.

\begin{figure}[t]
\begin{minipage}{.495\linewidth}
\centering
{\scriptsize $(p=5, n=500)$}\\
\vspace{1mm}
\includegraphics[keepaspectratio=true,width=.55\linewidth]{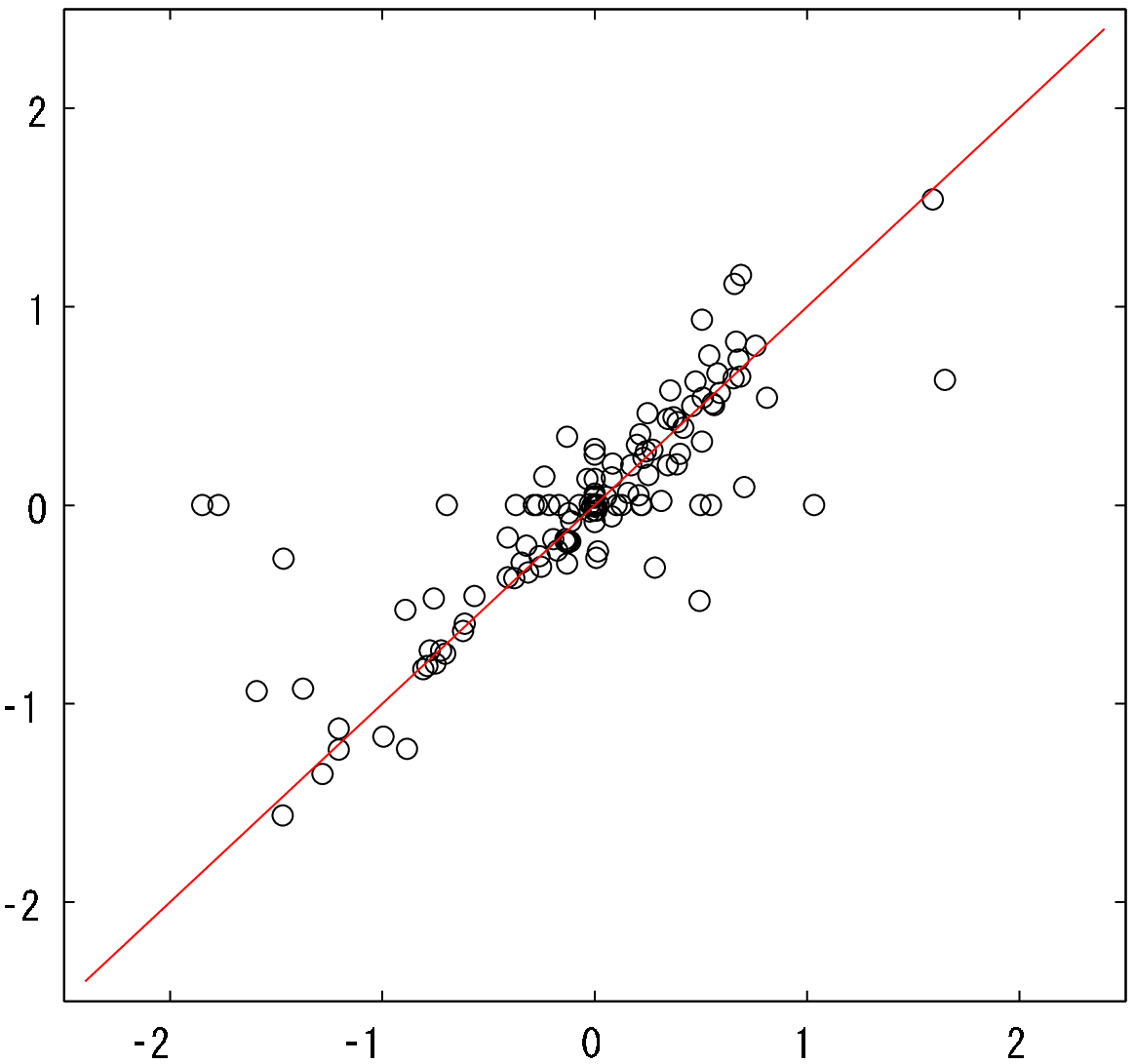}
\end{minipage}
\begin{minipage}{.495\linewidth}
\centering
{\scriptsize $(p=5, n=1000)$}\\
\vspace{1mm}
\includegraphics[keepaspectratio=true,width=.55\linewidth]{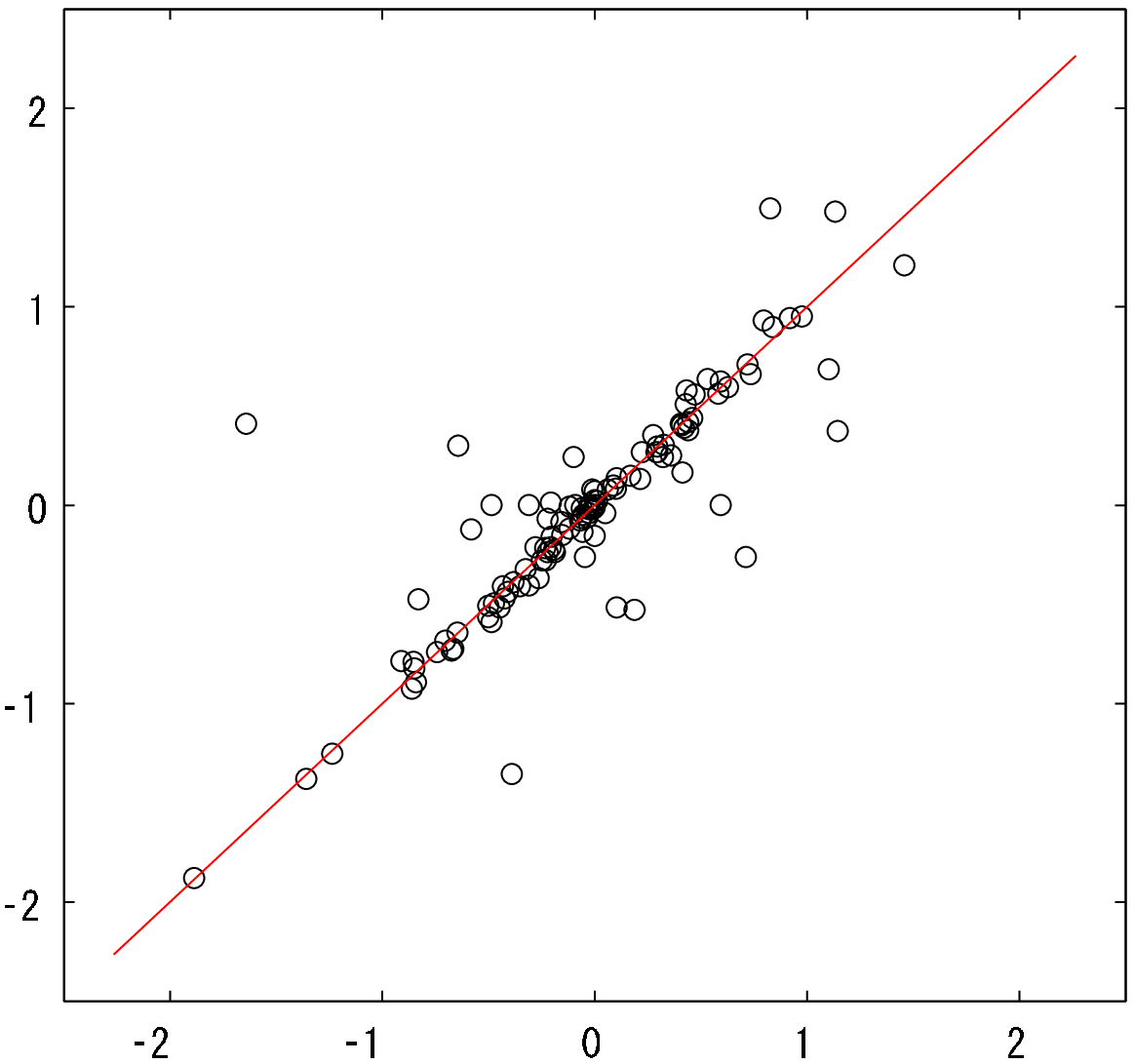}
\end{minipage}
\begin{minipage}{.495\linewidth}
\centering
\vspace{2mm}
{\scriptsize $(p=10, n=500)$}\\
\vspace{1mm}
\includegraphics[keepaspectratio=true,width=.55\linewidth]{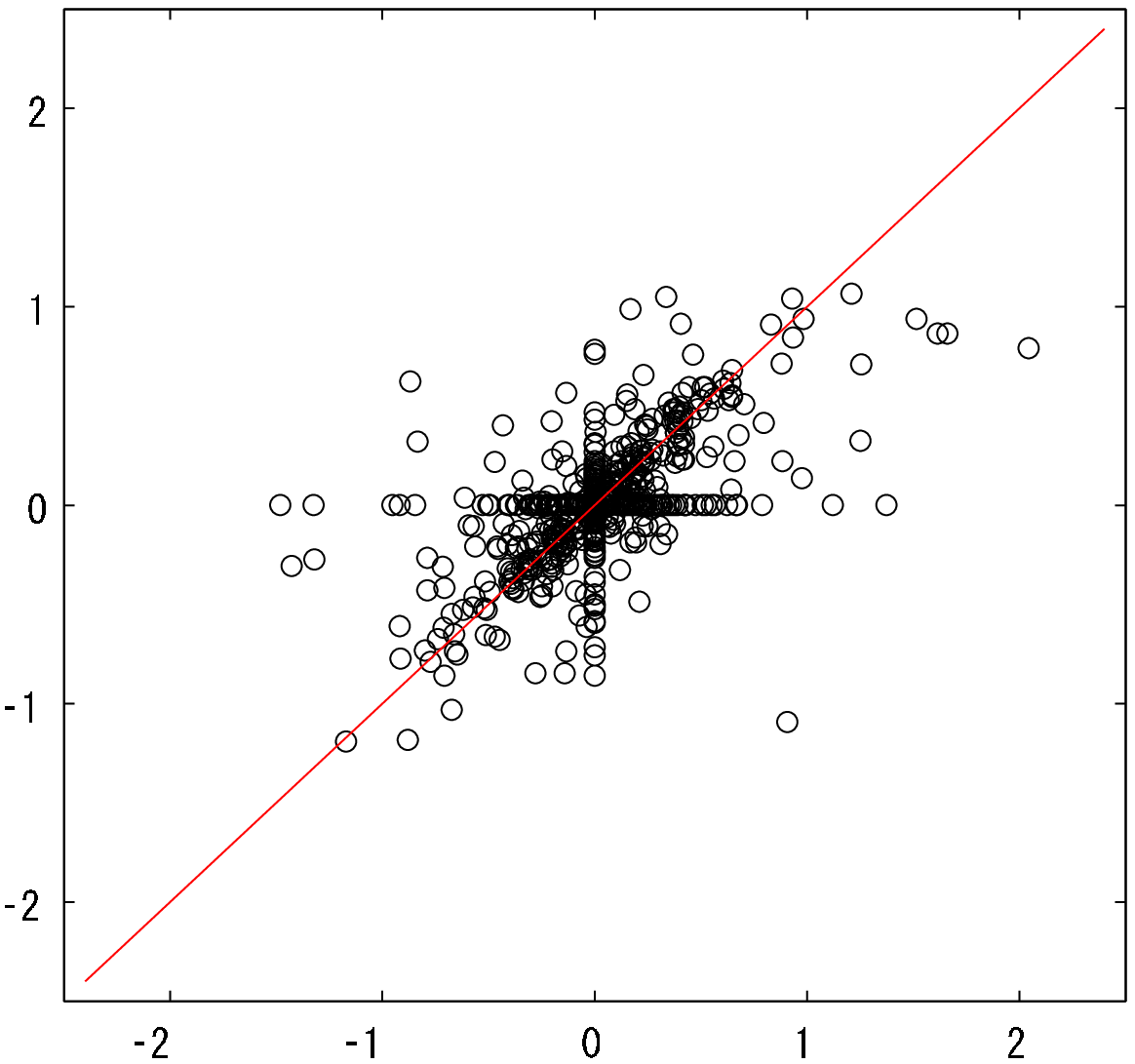}
\end{minipage}
\begin{minipage}{.495\linewidth}
\centering
\vspace{2mm}
{\scriptsize $(p=10, n=1000)$}\\
\vspace{1mm}
\includegraphics[keepaspectratio=true,width=.55\linewidth]{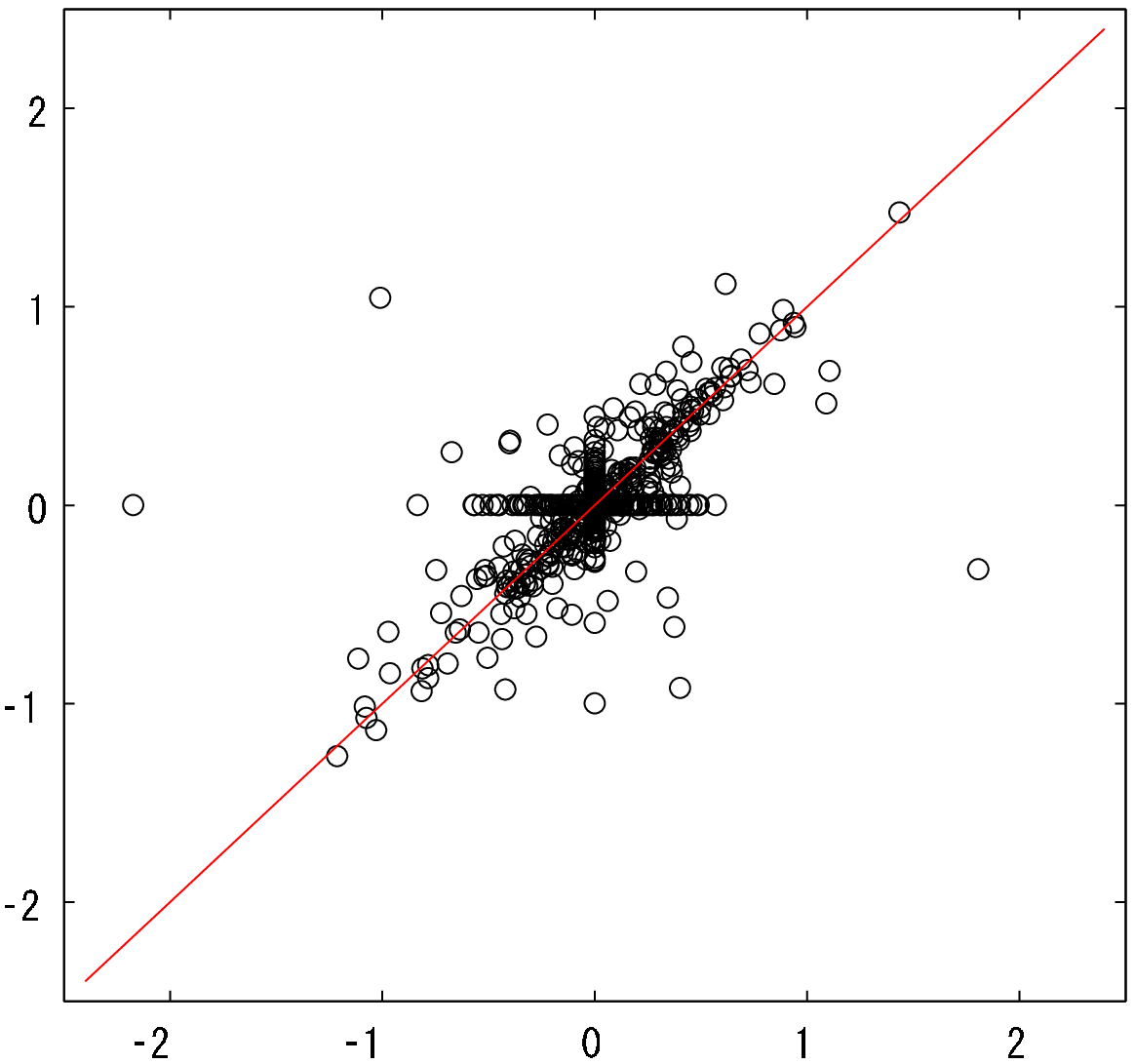}
\end{minipage}
\caption{Scatter-plots of the estimated $b_{ij}$ by Alg.~\ref{alg:glingam} (vertical axis) versus the generating values (horizontal axis) for combinations of dimensionality $p=(5, 10)$ and the number of samples $n=(500, 1000)$.}
\label{fig:scatter1}
\end{figure}

\begin{figure}[t]
\begin{minipage}{.495\linewidth}
\centering
{\scriptsize $(p=50, h=5)$}\\
\vspace{1mm}
\includegraphics[keepaspectratio=true,width=.55\linewidth]{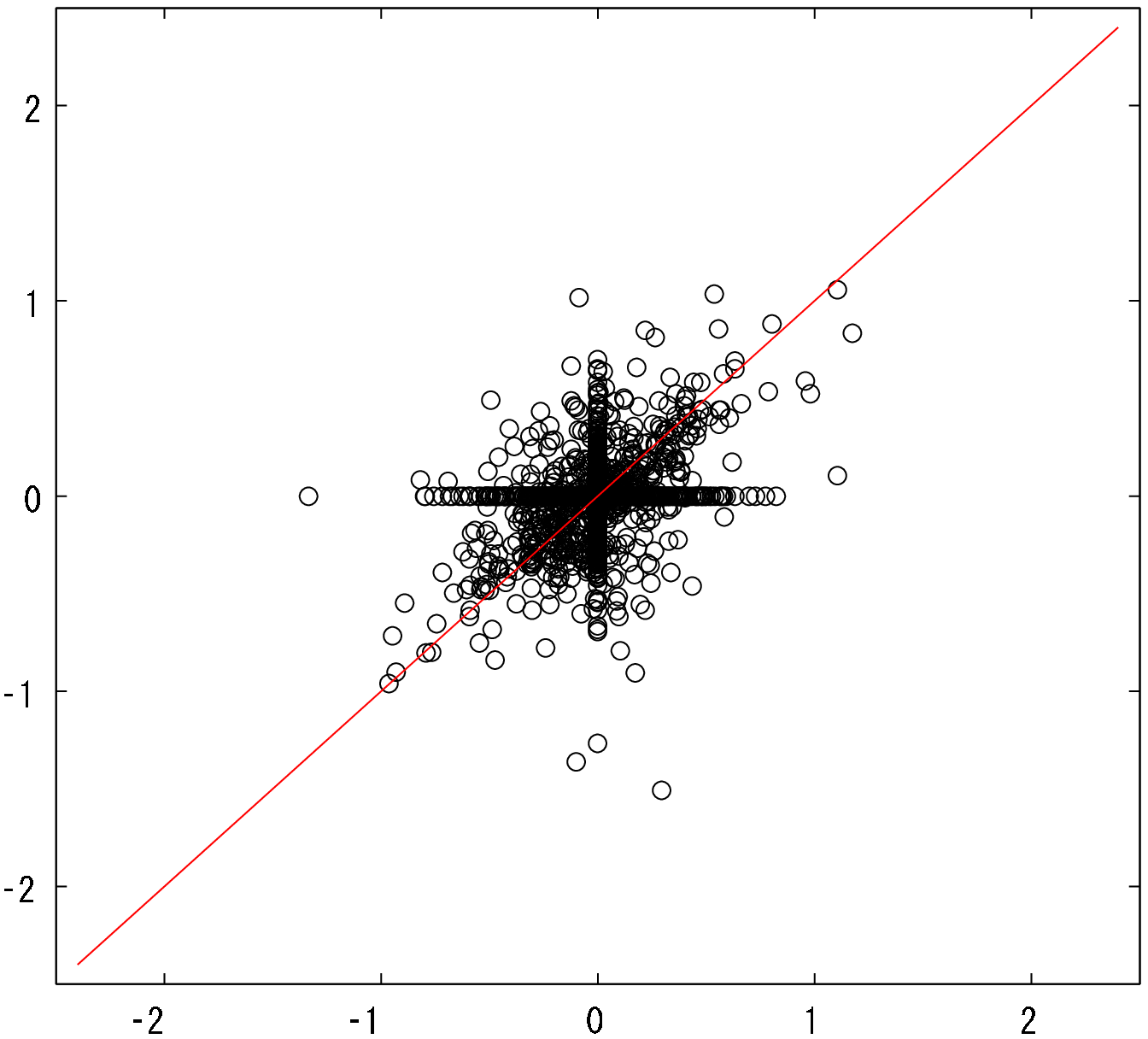}
\end{minipage}
\begin{minipage}{.495\linewidth}
\centering
{\scriptsize $(p=50, h=8)$}\\
\vspace{1mm}
\includegraphics[keepaspectratio=true,width=.55\linewidth]{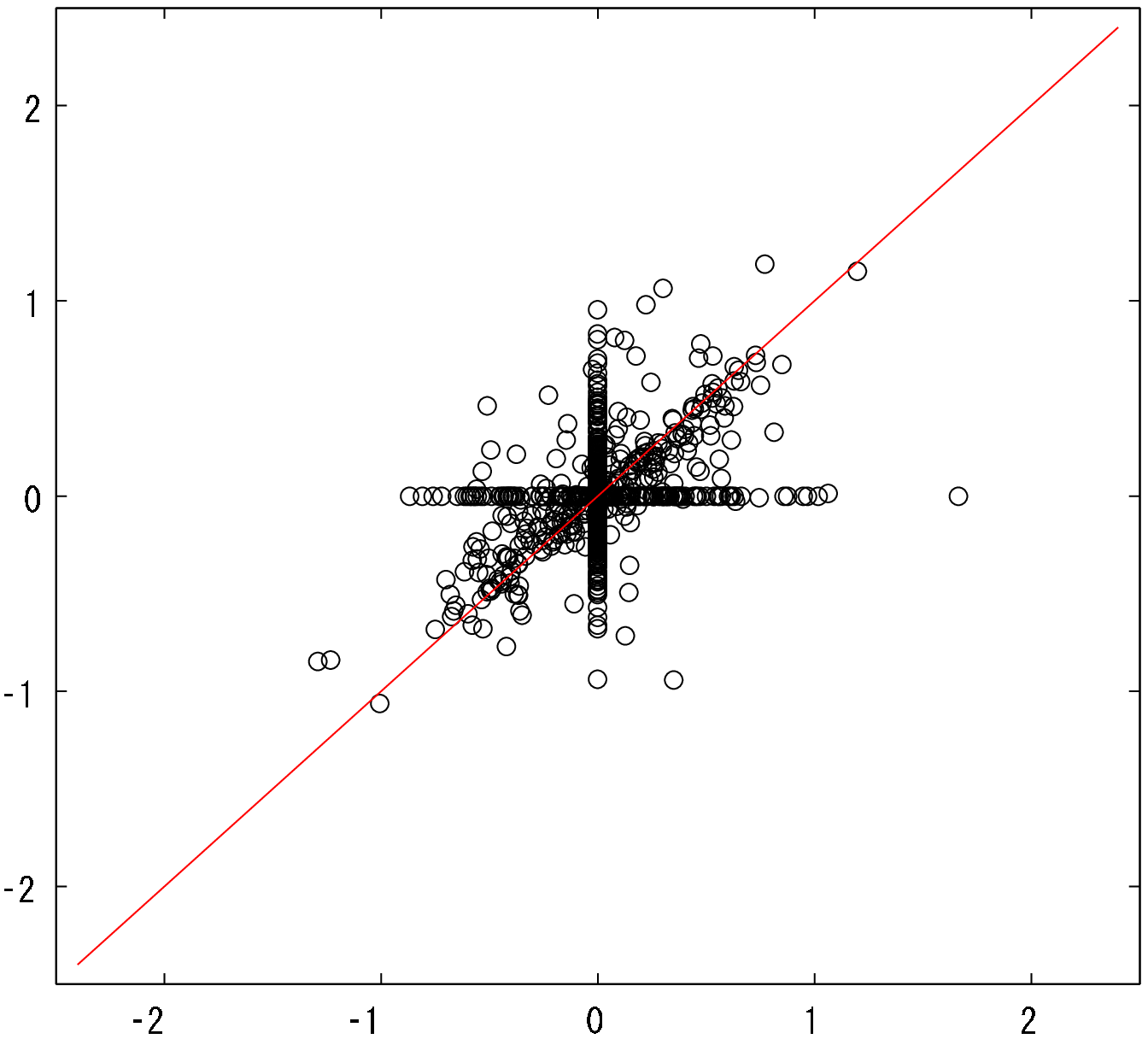}
\end{minipage}
\begin{minipage}{.495\linewidth}
\centering
\vspace{2mm}
{\scriptsize $(p=100, h=5)$}\\
\vspace{1mm}
\includegraphics[keepaspectratio=true,width=.55\linewidth]{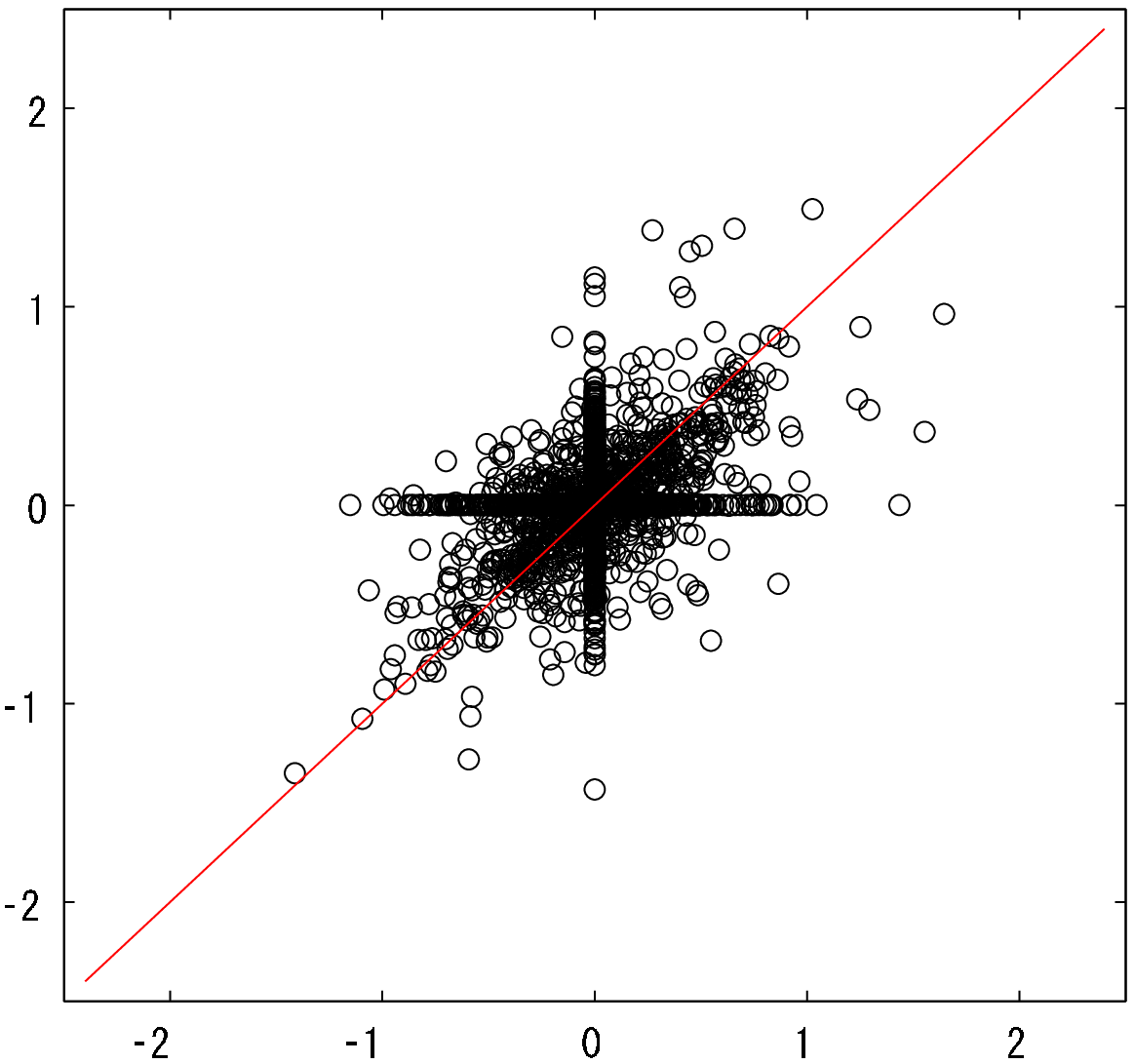}
\end{minipage}
\begin{minipage}{.495\linewidth}
\centering
\vspace{2mm}
{\scriptsize $(p=100, h=8)$}\\
\vspace{1mm}
\includegraphics[keepaspectratio=true,width=.55\linewidth]{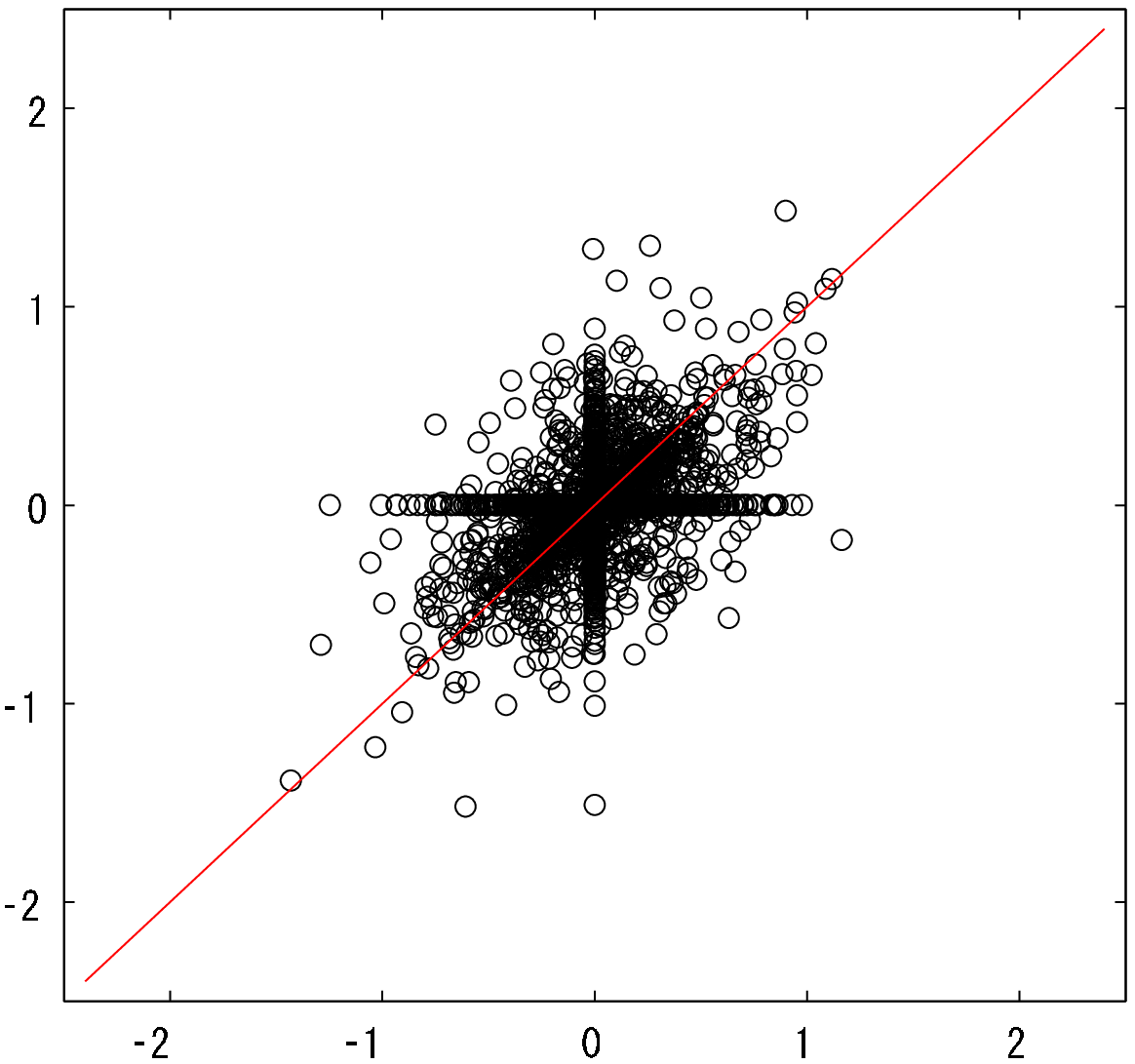}
\end{minipage}
\caption{Scatter-plots of the estimated $b_{ij}$ by Alg.~\ref{alg:glingam2} (vertical axis) versus the generating values (horizontal axis) for combinations of dimensionality $p=(50,$ $100)$ and coverage cardinality $h=(5, 8)$ ($n=1000$).}
\label{fig:scatter2}
\end{figure}

First, for the purpose (i), we created datasets under each combination of number of variables $p$, sample size $n$ and coverage cardinality $h$ (for Alg.~\ref{alg:glingam2}), as follows.\footnote{The way of creating datasets is the same as \cite{SHKW09} except for that $B$ is a block lower-triangular.}
\begin{enumerate}
\item First, a $p\times p$ block lower-triangular matrix $B$ was randomly created so that the standard deviations of variables owing to their parents (determined from its ordering of the subsets of variables) ranged in the interval $[0.5,1.5]$, where the number of blocks and the maximum number of parents of the created network for $B$ were also randomly determined from $1$ to $p$ in uniform manner. The standard deviations of the external influences $\boldsymbol{e}$ were randomly selected from the interval $[0.5,1.5]$.
\item Next, we generated data with sample size $n$ by independently drawing the external influence variables $\boldsymbol{e}$ from various non-Gaussian distributions with zero mean and unit variance. This is performed by generating Gaussian variables $z_i$ with zero means and unit variances, transformed it as $e_i=\text{sign}(z_i)|z_i|^{q_i}$, where nonlinear exponents $q_i$ were randomly selected from the interval $[0.5,0.8]\cup[1.2,2.0]$,\footnote{Nonlinear exponents $q_i$ with $[0.5,0.8]$ and $[1.2,2.0]$ give sub-Gaussian and super-Gaussian variables, respectively.} and then standardizing $e_i$ to have zero means and unit variables.
\item The values of the observed variables $\boldsymbol{x}$ were generated according to the GroupLiNGAM model \eqref{eq:glingam2}. And, the order of $\boldsymbol{x}$ is permuted randomly.
\end{enumerate}
The graphs in Fig.~\ref{fig:scatter1} and Fig.~\ref{fig:scatter2} show the scatter-plots of the elements of the estimated and generating adjacency matrix $B$ (for randomly generated 10 datasets in the respective case). GroupLiNGAM (Alg.~\ref{alg:glingam}) and L-GroupLiNGAM (Alg.~\ref{alg:glingam2}) were respectively applied~to relatively small and large sized graphs ($p$$=$$5,10$ (Fig.~\ref{fig:scatter1}) and $p$$=$$50,100$ (Fig.~\ref{fig:scatter2})). The parameters $\delta$ and $kneig$ were set as $1.0\times 10^{-2}$ and $0.05\times n$, respectively. For Alg.~\ref{alg:glingam2}, the number of subsets in a covering, {\em i.e.}, $N$, was set as $50$ in the experiment. Although the estimation seems to fail sometimes depending on dimensionality $p$, the number of samples $n$ or coverage cardinality $h$, the estimation seems to work reasonably well.

\begin{figure}[t]
\begin{minipage}{.495\linewidth}
\centering
\includegraphics[keepaspectratio=true,width=.72\linewidth]{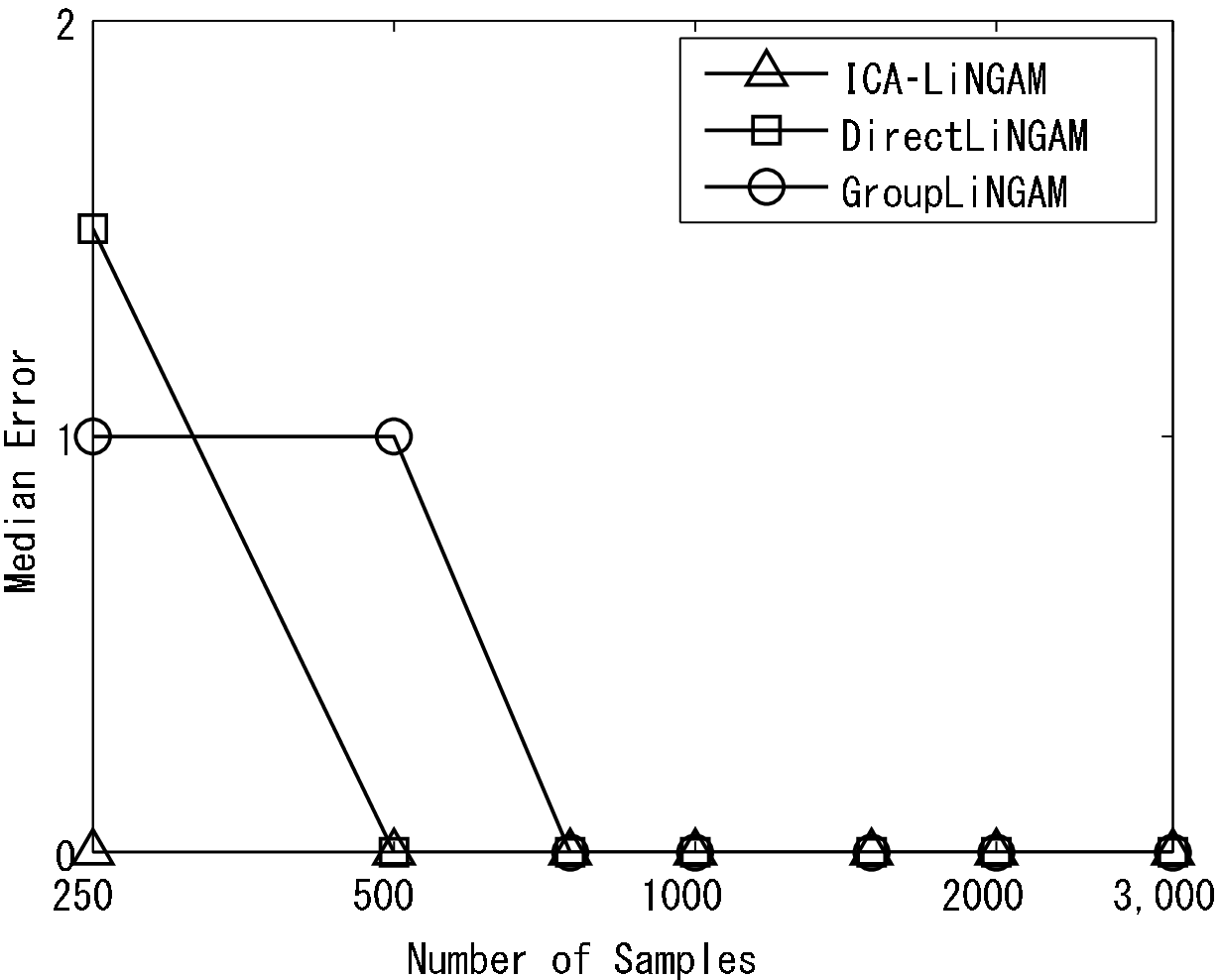}
\end{minipage}
\begin{minipage}{.495\linewidth}
\centering
\includegraphics[keepaspectratio=true,width=.72\linewidth]{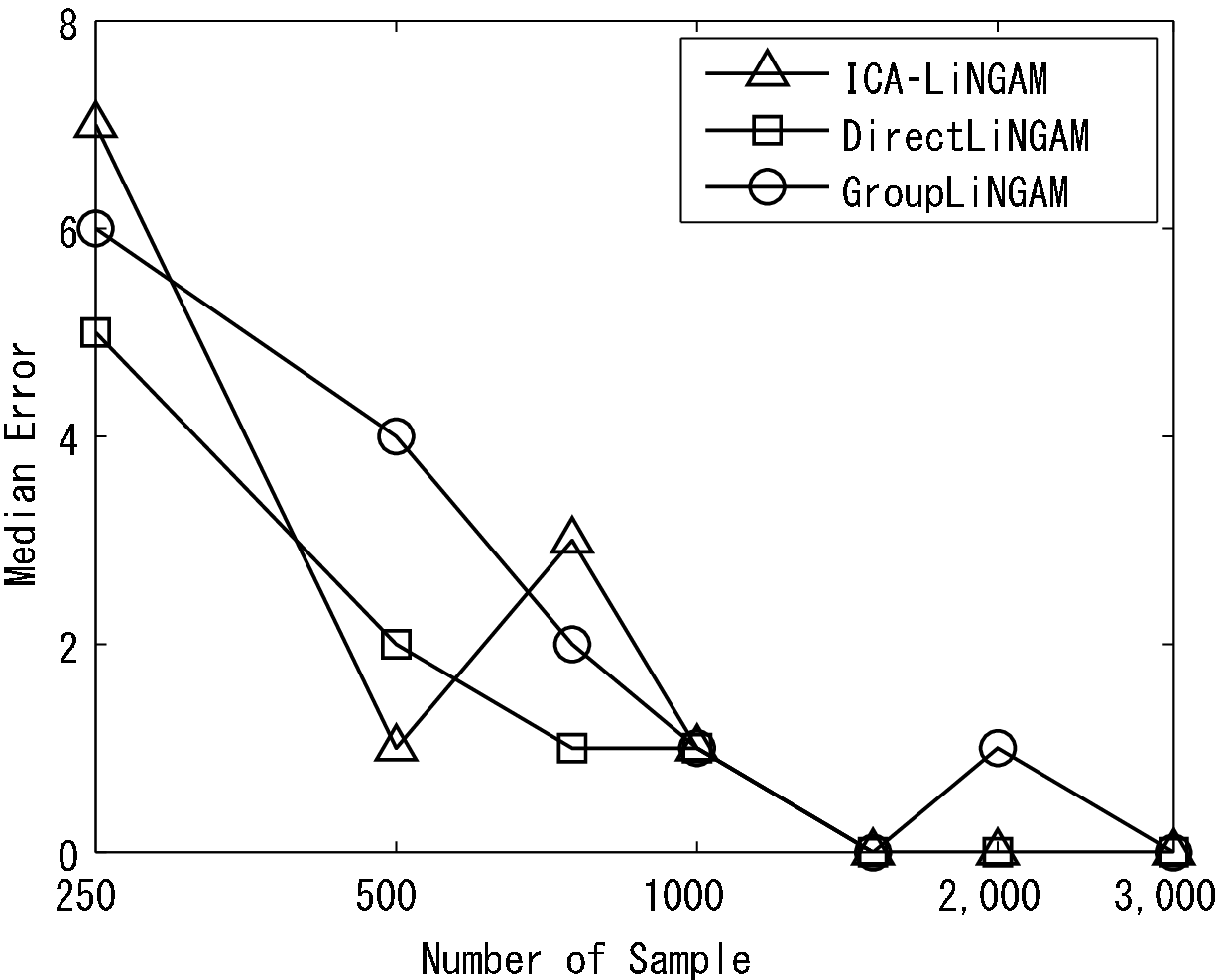}
\end{minipage}
\caption{Median numbers of errors in estimated orders by the existing and proposed algorithms when applied to DAG cases (Left: $p=6$ and Right: $p=10$).}
\label{fig:errors}
\end{figure}

Next, for the purpose (ii), we created datasets as in the same manner with the procedure described in \cite{SHHK06}, which is same with the above procedure except that each block contains only one element. As described above, GroupLiNGAM is specialized to the DAG case by setting $\delta$ as $+\infty$ and thus, in this experiment, $\delta$ was set as $1\times 10^6$ for Alg.~\ref{alg:glingam}. The graphs in Fig.~\ref{fig:errors} show the medians of the numbers of errors, {\em i.e.}, the numbers of elements in the strictly upper triangular part when the {\em true} connection strength matrix $B$ is permuted according to the estimated orders by the algorithms. The median errors by the algorithms are similar for almost experimental conditions and, hence, we can say that the estimation of GroupLiNGAM works reasonably well in the DAG case too. Here, we should note again that GroupLiNGAM can be applied not only to DAGs but also to chain graphs while the existing algorithms (ICA-LiNGAM and DirectLiNGAM) cannot.

\section{Application to real data}
\label{se:real}

\begin{figure}[t]
\centering
\includegraphics[keepaspectratio=true,width=.55\linewidth]{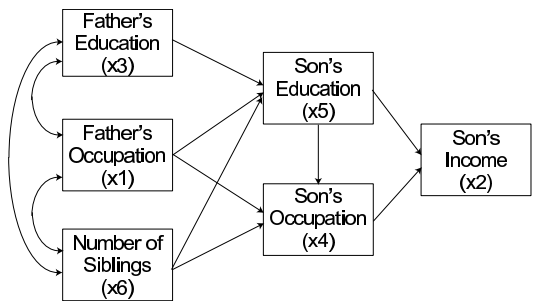}
\caption{Status attainment model based on domain knowledge, where $\{1,3,6\}$$<$$\{5\}$$<$$\{4\}$$<$$\{2\}$.}
\label{fig:background}
\end{figure}

To evaluate the applicability of the proposed algorithm (GroupLiNGAM), we analyzed a dataset taken from a sociological data repository on the Internet called General Social Survey.\footnote{\texttt{http://www.norc.org/GSS+Website/}}
The data consisted of six observed variables, $x_1$: father's occupation level, $x_2$: son's income, $x_3$: father's education, $x_4$: son's occupation level, $x_5$: son's education, $x_6$: number of siblings. 
The sample size was 1,380. Fig.~\ref{fig:background} shows domain knowledge about their causal relations: $K(1)$$=$$\{1,3,6\}$, $K(2)$$=$$\{5\}$, $K(3)$$=$$\{4\}$ and $K(4)$$=$$\{2\}$. 
In this section, we represent such relations by $\{1,3,6\}$$<$$\{5\}$$<$$\{4\}$$<$$\{2\}$ to save space. 
Note that if $\{i,j\}$$<$$\{k\}$, $x_i$ and $x_j$ could directly and/or indirectly cause $x_k$, but not vice versa.

In this experiment, Alg.~\ref{alg:glingam} was applied since the number of variables is small.
We tested several numbers of nearest neighbors $kneig=40,50,60,70$ to compute mutual information using the $k$-nearest neighbor approach \cite{KSG04} for GroupLiNGAM. 
The estimated networks were not sensitive to the choice of the number of nearest neighbors $kneig$, and essentially the same results were obtained for the values of $kneig$. 
We show the results under $kneig$$=$50 in Tab.~\ref{tab:est}, where a smaller threshold value for independence $\delta$ gives a network between larger groups of variables. 

We first analyzed all the six variables. 
The estimated orders by ICA-LiNGAM \cite{SHHK06}, Direct-LiNGAM \cite{SHKW09} and GroupLiNGAM are shown at the second top of Tab.~\ref{tab:est}. 
Those estimated orders are difficult to interpret since son's income ($x_2$) and/or son's education ($x_5$) could cause father's vari-
\begin{table}[!h]
\begin{center}
\begin{tabular}{ll}
Domain knowledge: & $\{1,3,6\}$$<$$\{5\}$$<$$\{4\}$$<$$\{2\}$\\
\hline
\multicolumn{2}{l}{All the six variables analyzed.} \\
ICA-LiNGAM: & $\{5\}$$<$$\{6\}$$<$$\{3\}$$<$$\{1\}$$<$$\{4\}$$<$$\{2\}$\\
DirectLiNGAM: & $\{6\}$$<$$\{2\}$$<$$\{1\}$$<$$\{3\}$$<$$\{4\}$$<$$\{5\}$\\
GroupLiNGAM: & \\
\multicolumn{1}{r}{$\delta$$=$0.500} & $\{6\}$$<$$\{2\}$$<$$\{1\}$$<$$\{4\}$$<$$\{5\}$$<$$\{3\}$\\
\multicolumn{1}{r}{$\delta$$=$0.100} & $\{6\}$$<$$\{2\}$$<$$\{1\}$$<$$\{4,5\}$$<$$\{3\}$\\
\multicolumn{1}{r}{$\delta$$=$0.010} & $\{6\}$$<$$\{2\}$$<$$\{1,3,4,5\}$\\
\multicolumn{1}{r}{$\delta$$=$0.001} & $\{1,2,3,4,5,6\}$\\
\hline
\multicolumn{2}{l}{$x_2$ omitted.} \\
ICA-LiNGAM: & $\{5\}$$<$$\{6\}$$<$$\{3\}$$<$$\{1\}$$<$$\{4\}$\\
DirectLiNGAM: & $\{6\}$$<$$\{1\}$$<$$\{3\}$$<$$\{4\}$$<$$\{5\}$\\
GroupLiNGAM: & \\
\multicolumn{1}{r}{$\delta$$=$0.500} & $\{6\}$$<$$\{1\}$$<$$\{3\}$$<$$\{5\}$$<$$\{4\}$\\
\multicolumn{1}{r}{$\delta$$=$0.100} & $\{6\}$$<$$\{1,3\}$$<$$\{5\}$$<$$\{4\}$\\
\multicolumn{1}{r}{$\delta$$=$0.050} & $\{6\}$$<$$\{1,3\}$$<$$\{4,5\}$\\
\multicolumn{1}{r}{$\delta$$=$0.010} & $\{6\}$$<$$\{1,3,4,5\}$\\
\multicolumn{1}{r}{$\delta$$=$0.001} & $\{1,3,4,5,6\}$\\
\hline
\multicolumn{2}{l}{$x_2$ and $x_6$ omitted.} \\
ICA-LiNGAM: & $\{5\}$$<$$\{3\}$$<$$\{1\}$$<$$\{4\}$\\
DirectLiNGAM: & $\{1\}$$<$$\{3\}$$<$$\{4\}$$<$$\{5\}$\\
GroupLiNGAM: & \\
\multicolumn{1}{r}{$\delta$$=$0.50} & $\{3\}$$<$$\{1\}$$<$$\{5\}$$<$$\{4\}$\\
\multicolumn{1}{r}{$\delta$$=$0.10} & $\{1,3\}$$<$$\{5\}$$<$$\{4\}$\\
\multicolumn{1}{r}{$\delta$$=$0.05} & $\{1,3\}$$<$$\{4,5\}$\\
\multicolumn{1}{r}{$\delta$$=$0.01} & $\{1,3,4,5\}$$$
\end{tabular}
\end{center}
\caption{Estimated orders of groups.}
\label{tab:est}
\end{table}

ables ($x_1$,$x_3$), but not vice versa. The orders are not reasonable to their temporal orderings. 

Next, we omitted son's income ($x_2$) and analyzed the other five variables. Omitting $x_2$ would not create any unobserved confounder since it does not cause any other variables according to the domain knowledge. 
The results are shown in the third top of Tab.~\ref{tab:est}. 
DirectLiNGAM and GroupLiNGAM found consistent time orderings between father's variables ($x_1$,$x_3$) and son's variables ($x_4$,$x_5$). 
Furthermore, GroupLiNGAM found a reasonable ordering between son's variables, {\it i.e.}, son's education ($x_5$) could cause son's occupation level ($x_4$), but not vice versa, whereas DirectLiNGAM failed. 
However, number of siblings ($x_6$) is the top variable in every estimated ordering by DirectLiNGAM and GroupLiNGAM and could cause father's variables ($x_1$,$x_3$), which is not easy to interpret. 

We further omitted number of siblings ($x_6$) as well as son's income ($x_2$) and analyzed the other four variables ($x_1$,$x_3$,$x_4$,$x_5$). 
Omitting $x_6$ could create an unobserved confounder since it could relate father's variables ($x_1$,$x_3$) and son's variables ($x_4$,$x_5$). 
The bottom of Tab.~\ref{tab:est} shows the results. 
Every ordering estimated by GroupLiNGAM is consistent with the domain knowledge. 
ICA-LiNGAM wrongly estimated that son's education ($x_5$) could cause father's variables ($x_1$,$x_3$), but not vice versa. DirectLiNGAM also gave inconsistent orderings between father's education ($x_3$) and father's occupation ($x_1$) and between son's education ($x_5$) and son's occupation ($x_4$). 

In summary, GroupLiNGAM provided more consistent orderings with the domain knowledge than ICA-LiNGAM and DirectLiNGAM. The reason would be that only GroupLiNGAM is able to allow unobserved confounders. 
However, it is not yet very clear why the inclusion of $x_2$ and $x_6$ makes the results difficult to interpret. 
One possibility is that $x_2$ and $x_6$ might not fit well some assumption in the three discovery methods, {\em e.g.}, linearity,  compared to the other four variables. 

\section{Conclusions}
\label{se:conclusion}

In this paper, we proposed the GroupLiNGAM model, a non-Gaussian variant of chain graphs, and presented an algorithm for estimating this model, which is identifiable without any prior knowledge on the structure. Based on the result that an exogeneous set is identified by evaluating the independence between a variable subset and the residuals when the remaining variables are regressed on those, the proposed algorithm finds an ordered devision of variables iteratively and identifies an ordering of disjoint subsets of variables. However, since the computational cost grows exponentially according to the number of variables, a middle sized graph is the practical limit of this algorithm. Therefore, in addition, we presented an approximate approach to apply this framework to large sized graphs. In the experimental parts, we evaluated the algorithms empirically and illustrated the applicability using artificial and real datasets.

The algorithm has a tuning parameter $\delta$, which determines when the devision of groups should be stopped ($kneig$ is also an tuning parameter in the current implementation. However, this parameter is for the estimation of mutual information by the $k$-nearest neighbor method \cite{KSG04} and thus would not be an essential parameter in our method). For more exact devision of groups, it would be useful to combine our framework with some statistical test method, such as the bootstrap method \cite{ET94}, in the future. Also, in the current implementation, exponentially large number of subsets need to be examined when identifying an exogenous (Line1--3 in GroupSearch in Alg.~\ref{alg:glingam}). Therefore, it would be important to develop more efficient search strategy for this part using some discrete structure.

{\small
\bibliography{gLiNGAM_arxiv}
\bibliographystyle{amsplain}}
 
\end{document}